\documentclass[english]{article}

\usepackage[colorlinks,bookmarksopen,bookmarksnumbered,allcolors=red]{hyperref}
\usepackage[table, dvipsnames]{xcolor}
\usepackage{graphicx}
\usepackage{geometry}
\usepackage{amsmath, amssymb, bm}
\usepackage{relsize}
\usepackage{authblk}
\usepackage{setspace}
\usepackage{dsfont}
\usepackage{listings}
\usepackage{float} 
\usepackage{multirow}

\definecolor{bluegreen}{RGB}{46,141,131}
\definecolor{darkgreen}{RGB}{46,139,87}
\definecolor{darkred}{RGB}{219,7,61}
\definecolor{darkblue}{RGB}{0,0,137}

\title{\textbf{The LongiMam model for improved breast cancer risk prediction using longitudinal mammograms}}

\author[,1]{Manel Rakez\thanks{Corresponding author: manel.rakez@u-bordeaux.fr}}
\author[2]{Thomas Louis}
\author[2]{Julien Guillaumin}
\author[3]{Foucauld Chamming's}
\author[2]{Pierre Fillard}
\author[$\dagger$,4]{Brice Amadeo}
\author[,1]{Virginie Rondeau\thanks{Brice Amadeo and Virginie Rondeau contributed equally as co-last authors}}
\affil[1]{BIOSTAT team, Bordeaux Population Health U1219, Bordeaux University, ISPED, Bordeaux, France}
\affil[2]{Therapixel, Nice, France}
\affil[3]{Department of Radiology, Institut Bergonié, Comprehensive Cancer Centre, Bordeaux, France}
\affil[4]{EPICENE team, Bordeaux Population Health U1219, Bordeaux University, ISPED, Bordeaux, France}

\date{}

\begin{document}

\maketitle

\begin{abstract}
Risk-adapted breast cancer screening requires robust models that leverage longitudinal imaging data. Most current deep learning models use single or limited prior mammograms and lack adaptation for real-world settings marked by imbalanced outcome distribution and heterogeneous follow-up. We developed LongiMam, an end-to-end deep learning model that integrates both current and up to four prior mammograms. LongiMam combines a convolutional and a recurrent neural network to capture spatial and temporal patterns predictive of breast cancer. The model was trained and evaluated using a large, population-based screening dataset with disproportionate case-to-control ratio typical of clinical screening. Across several scenarios that varied in the number and composition of prior exams, LongiMam consistently improved prediction when prior mammograms were included. The addition of prior and current visits outperformed single-visit models, while priors alone performed less well, highlighting the importance of combining historical and recent information. Subgroup analyses confirmed the model's efficacy across key risk groups, including women with dense breasts and those aged 55 years or older. Moreover, the model performed best in women with observed changes in mammographic density over time. These findings demonstrate that longitudinal modeling enhances breast cancer prediction and support the use of repeated mammograms to refine risk stratification in screening programs. \href{https://github.com/manelrakez/LongiMam.git}{LongiMam} is publicly available as open-source software.\newline
{\emph{Keywords:}} Breast cancer risk, Digital mammography, Longitudinal data, Convolutional neural network, Gated recurrent unit.
\end{abstract}

\section{Introduction}
\label{sec::intro}

In the United States, breast cancer is the most frequently diagnosed cancer in women, with an estimated $42{,}250$ deaths expected in 2024 \cite{american_cancer_society_breast_nodate}. Because breast cancer often has no symptoms at an early, more treatable stage, mammography screening plays a central role in early detection. Since 1990, screening combined with therapeutic advances has contributed to a 41\% reduction in breast cancer mortality \cite{rahman_breast_2022}.

Current screening protocols are based on fixed schedules and broad risk categories. In the US, guidelines recommend annual mammography for women at average risk starting at age 40, with emphasis on regular screening from age 45 \cite{oeffinger_breast_2015}. Each exam consists of two images per breast in two views (craniocaudal (CC) and mediolateral oblique (MLO)), resulting in four images per visit. These exams are repeated at regular intervals until a diagnosis or an upper age limit, creating a sequence of mammograms over time. While effective at the population level, this approach can lead to overdiagnosis and overtreatment in women at low risk if intervals are too short, and to missed cancers in high-risk women if intervals are too long \cite{patel_assessment_2021}. More precise, individualized risk models that leverage longitudinal mammography data are therefore of interest.

Breast cancer is a progressive disease, and sequential mammograms capture structural changes in breast tissue that may indicate risk. Several studies have shown that prior mammograms provide valuable information, improving cancer detection, reducing false recalls, and lowering interval cancer rates \cite{hayward_improving_2016, choi_analysis_2016, hovda_radiological_2021}. Cross-sectional models relying only on the most recent exam may therefore be insufficient. Radiologists routinely review both current and historical images, and predictive models should reflect this practice to encourage clinical implementation and adoption as a computed-aided diagnostic tool.

Several image-based deep learning methods have shown promise for breast cancer prediction and suggest that these models can extract more predictive information from mammograms than traditional risk models \cite{lee_enhancing_2023, yala_toward_2021, donnelly_asymmirai_2024}. Most of these approaches focus on 1- to 5-year risk prediction at the individual level and rely mainly on transformer-based architectures, sometimes combined with available or imputed clinical characteristics. While such long-term prediction can provide useful additional information for radiologists once the models are validated for clinical practice, current implementations make limited use of a woman’s screening history. In most cases, they rely on no or very few prior examinations, which reduces their potential to capture longitudinal information. Moreover, risk estimation is often based on additive hazard formulations that compute yearly risks in parallel without conditioning on survival up to each time point, raising concerns about the validity of the estimated risks.

Another line of research has explored latent ordinary differential equations coupled with recurrent neural networks (ODE-RNNs) to address challenges in medical data, such as irregular and sparse observation times \cite{rubanova_latent_2019, moon_survlatent_2022}. These methods are attractive because follow-up schedules in healthcare often vary and missing visits are common. However, to our knowledge, they have not yet been applied to imaging data.

A recent advancement, the Longitudinal Risk Prediction NETwork (LRPNET) by Dadsetan et al. (2022) \cite{dadsetan_deep_2022}, aimed to capture spatiotemporal changes in breast tissue across serial mammograms. Integrating four prior visits produced better predictions compared to models that used fewer priors, simple averaging, or a single time point. However, LRPNET was trained and evaluated on a limited, balanced dataset of 200 women, unlike actual screening cohorts where cancer incidence is far lower. The model also froze the convolutional backbone (a pre-trained VGG16), which could restrict adaptability to heterogeneous data. Furthermore, LRPNET required exactly four prior exams for each subject and only incorporated negative mammograms from prior visits, omitting the current exam. These constraints limit the real-world utility and generalizability of the model.

In this study, we extend LRPNET to a large, population-based screening dataset with substantial imbalance in outcome distribution and introduce the LongiMam (Longitudinal Mammogram-based) model, an enhanced deep learning solution. LongiMam combines a convolutional neural network (CNN) and a recurrent neural network (RNN) to overcome LRPNET’s limitations in adaptability, data availability, and clinical applicability. We assess the model in several scenarios that differ in input sequence composition and length to determine the value of prior mammograms for breast cancer detection. The implementation is available as open-source software on \href{https://github.com/manelrakez/LongiMam.git}{GitHub}.

The remainder of this article is structured as follows: Section 2 describes the dataset, model architecture, and training methods. Section 3 presents model performance on real-world data in addition to some subgroup analyses. Section 4 discusses the findings and the broader implications for breast cancer risk prediction.

\section{Methods}
\label{sec::methods}

\subsection{Dataset}
\label{sec::met_dataset}
\subsubsection{Image selection}
\label{sec::image_selection}
This study utilized an in-house dataset curated and maintained by Therapixel, a French company specializing in artificial intelligence (AI) for medical imaging \cite{therapixel_artificial_2023}. The study is exempt from Subpart A of the federal regulations for the protection of human subjects (45 CFR 46), as it involves retrospectively collected, de-identified data and poses minimal risk.

\begin{figure}[htbp]
 \centerline{\includegraphics[width=6in]{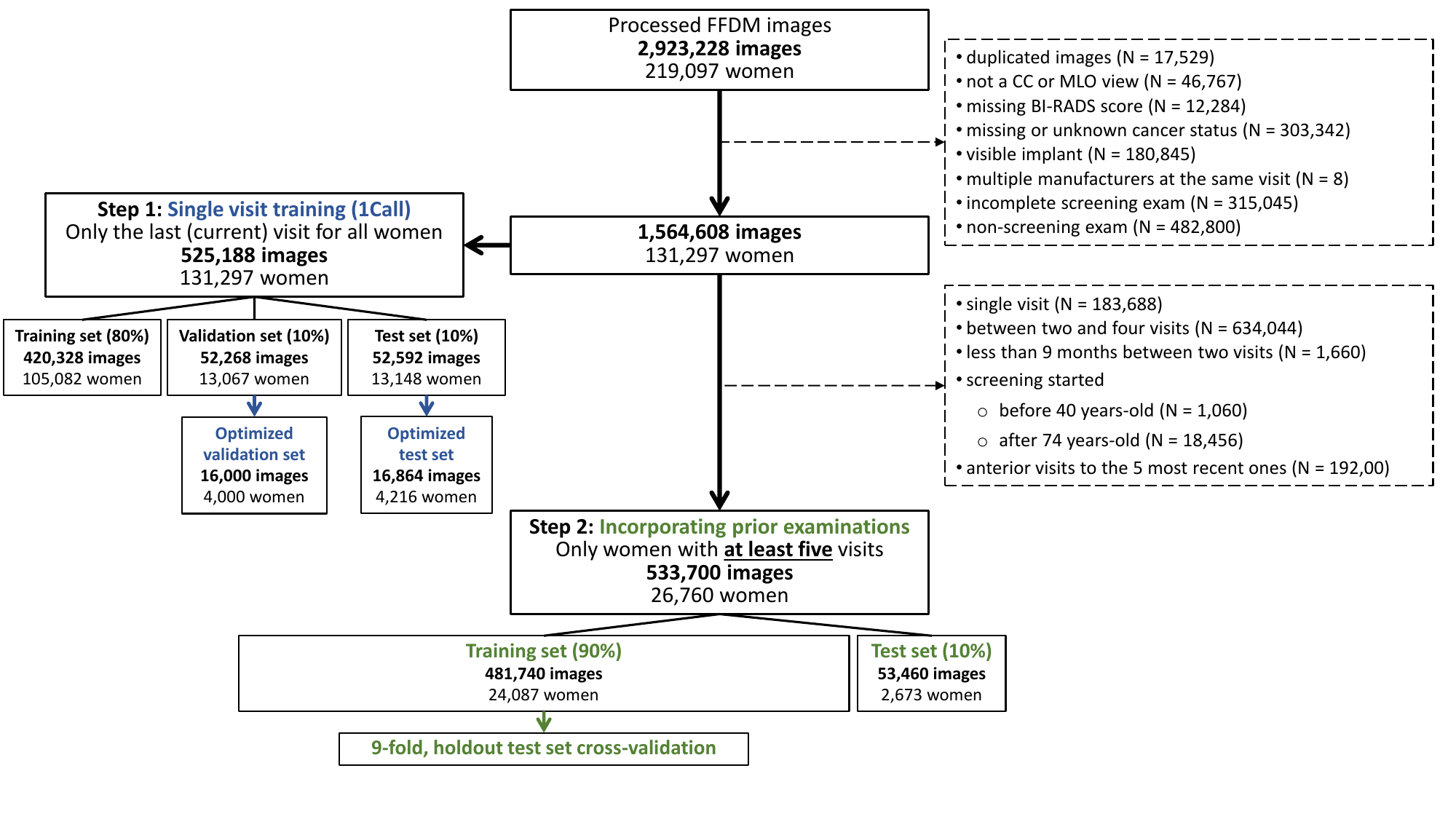}}
\caption{\textbf{Data selection flowchart}. After applying exclusion criteria, the dataset comprised 1,564,608 images from 131,297 women. For single-visit training (Step 1), only the most recent exam per woman was used. For longitudinal analyses (Step 2), only women with at least five visits were included, resulting in 533,700 images from 26,760 women. Both subsets were split into training, validation and test sets according to the analysis design.}
\label{f:flowchart}
\end{figure}

Data included retrospectively collected processed full-field digital mammography (FFDM) images from women screened between October 2006 and August 2019 in the United States, following a yearly screening schedule. All mammographic examinations were acquired by either Hologic Inc. or General Electric Healthcare (GEHC) mammography units. We excluded duplicate images, images captured in views other than CC and MLO, images with missing or unknown cancer status, images showing breast implants, examinations containing images from multiple manufacturers within the same visit, and incomplete or non-screening examinations. Data were sourced from two screening centers. Breast cancer status was confirmed by a positive biopsy following a suspicious mammography exam. A flowchart detailing the data selection process is provided in Figure~\ref{f:flowchart}.

For the second step of this work where we were interested in incorporating prior examinations, additional eligibility criteria were applied to ensure consistent longitudinal patterns. Specifically, we excluded women with fewer than five screening visits, visit intervals shorter than nine months, or screening initiation before age 40 or after age 74. For women with more than five visits, only the five most recent were retained.

Longitudinal information was structured as follows: For women diagnosed with breast cancer, the "current" visit denoted the exam showing the suspicious lesion that led to a positive biopsy. For controls, the "current" visit represented the last negative exam, confirmed by a subsequent screening visit with negative findings. Prior exams were indexed in reverse chronological order relative to the current visit: Prior 1 refers to the exam immediately preceding the current visit, Prior 2 to the second most recent, and so on, up to Prior 4. A minimum of 9 months was required between two visits to be considered distinct screening episodes. While most women adhered to annual screening schedules, we did not impose constraints on the time interval between visits in order to preserve real-world variability in screening behavior. No matching was performed between cases and controls. Each complete screening exam consisted of four images: two views (CC and MLO) for each breast.

\subsubsection{Image preprocessing}
\label{sec::image_preprocess}
To standardize image dimensions across the dataset, each original mammogram was resized along the height and either cropped or padded along the width to achieve a final resolution of 576×416 pixels. Pixel intensities were then rescaled to the [0,1] range using the Values of Interest Lookup Table (VOI LUT) transformation \cite{dicom_standard_browser_voi_nodate}. Background pixels were set to zero to facilitate the identification of relevant regions within the breast area.

To further improve the model's performance, data augmentation was performed at the breast side level for each subject \cite{buslaev_albumentations_2020}. Specifically, each mammogram was randomly altered by either a horizontal flip, a rotation (-10, 10 degrees), a shift (-0.05\%, 0.05\% in image height and width), or a change in brightness (-0.05\%, 0.05\%) and contrast (-0.1\%, 0.1\%). For consistency across time, the same randomly selected transformation was applied to all images from a given side, regardless of the timepoint.

\subsection{Model architechture}
\label{sec::met_model_compo}
\begin{figure}[htbp]
 \centerline{\includegraphics[width=6in]{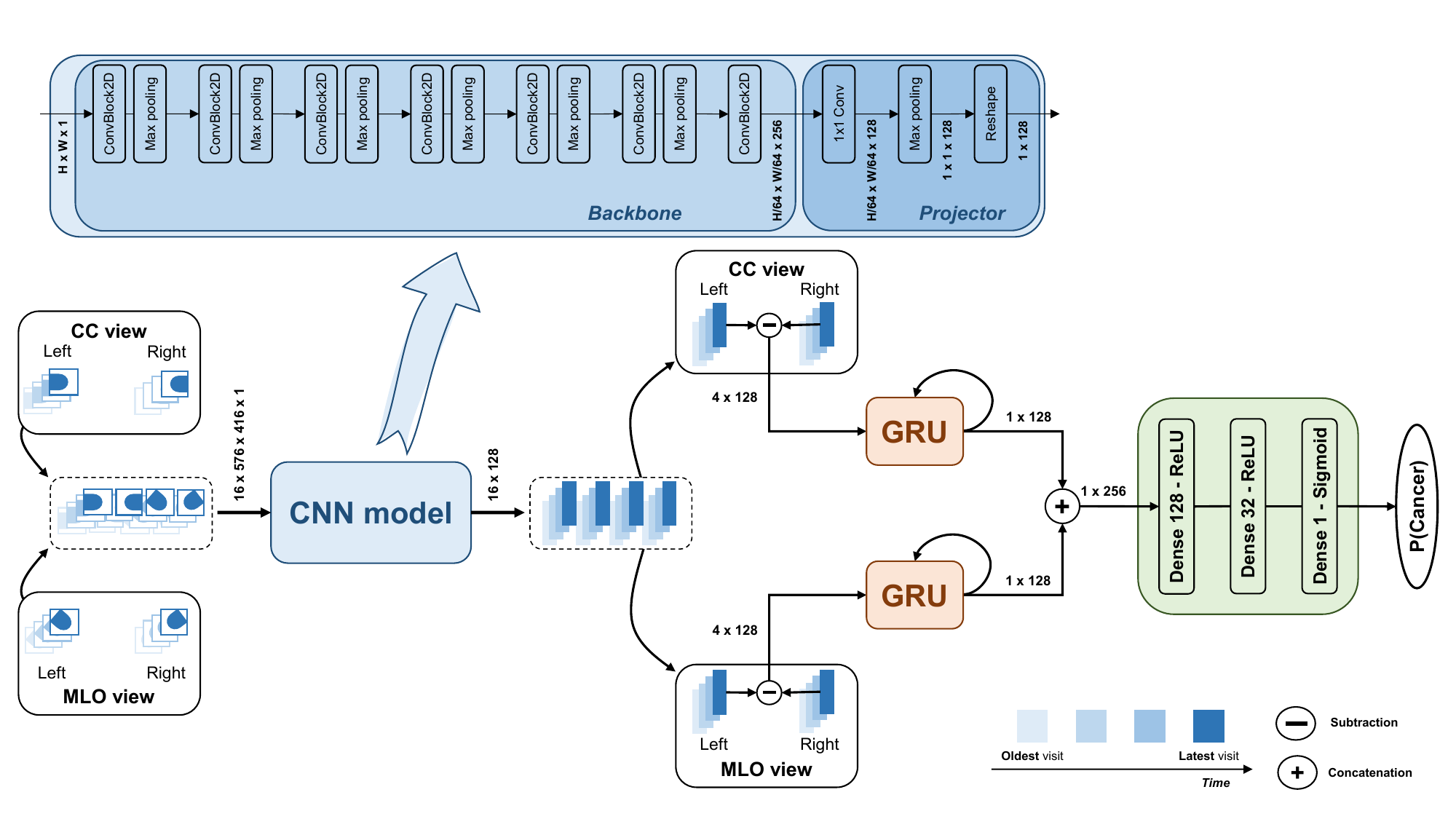}}
\caption{\textbf{Overview of the proposed longitudinal deep learning model for breast cancer risk prediction.} Each exam comprises four grayscale images: craniocaudal (CC) and mediolateral oblique (MLO) views of the left and right breasts, across multiple sequential screening visits (Fours visits in this example). Images are processed by a convolutional neural network (CNN) backbone and projector to extract feature vectors for each view and timepoint. For each view, left-right differences across visits are computed and passed to gated recurrent units (GRUs) to model temporal patterns across visits. The outputs from the CC and MLO GRUs are concatenated and passed through fully connected layers with a final sigmoid layer to estimate the probability of cancer. Color intensity denotes time (lightest: oldest; darkest: most recent).}
\label{f:model_archi}
\end{figure}

As in Dadsetan et al. (2022) \cite{dadsetan_deep_2022}, our model consists of three main components: a convolutional neural network (CNN) model, a recurrent neural network (RNN)--specifically a gated recurrent unit (GRU)--and a final classification module composed of dense layers. Below, we detail the key modifications introduced in each component relative to the original LRPNET model. An overview of the architecture is presented in Figure~\ref{f:model_archi}.

\subsubsection{CNN model for feature extraction}
\label{sec::met_cnn_backbone}
In contrast to LRPNET, which uses a CNN model based on a pre-trained VGG16 backbone, we employed an in-house CNN architecture composed of two parts represented by a backbone and a projector. Our backbone comprised a repeated convolutional blocks (denoted ConvBlock) where each ConvBlock includes a 3×3 convolution, batch normalization, and a ReLU activation. This block is repeated six times, with each repetition followed by a 2×2 max-pooling layer. After the sixth ConvBlock, we apply a final ConvBlock to the resulting feature maps of size Height/64 x Width/64 x 256. Next, we define the projector composed by a 1×1 convolution to reduce the depth of the input vector to 128 channels, followed by a final max-pooling step to yield a feature map of size 1x1x128, which is then reshaped into a 1×128 feature vector.

This CNN architecture was designed to process all four mammograms composing the complete screening exam, regardless of view, laterality or time point. Training was performed with a batch defined at the subject level, with each woman contributing either single of multiple visits depending on the considered scenario (See Section~\ref{sec::met_scenarios}). To accelerate training, we initialized our backbone using a pre-trained version previously developed for lesion detection in mammograms, internally provided by Therapixel \cite{therapixel_artificial_2023}.

\subsubsection{GRU block for longitudinal information processing}
\label{sec::met_gru}
We employed a many-to-one GRU architecture, similar to the approach used by Dadsetan et al. (2022) \cite{dadsetan_deep_2022}. Separate GRU blocks were defined for the CC and MLO views. Each GRU receives as input the element-wise difference between the feature vectors of the left and right breasts for the corresponding view, computed at each time point. This subtraction aims to capture asymmetries between the breasts over time.\\
Each GRU outputs a vector of size 1×128. The outputs from the CC and MLO GRU blocks are then concatenated to form a single 1×256 vector, which is passed to the final classification block.

\subsubsection{Dense layers for classification}
\label{sec::met_fcl}
The classification module (in green in Figure~\ref{f:model_archi}) consists of three consecutive dense layers with 128, 32, and 1 node(s), respectively. A sigmoid activation function is applied to the final output to estimate the probability that a given subject will develop breast cancer based on her mammographic history. No dropout regularization was applied in this block.

\subsection{Training strategy}
\label{sec::met_training_strategy}
We adopted a two-step training strategy to improve model's performance and allow for swift training in the second step when priors are added.\\
In the first step, we trained the full model architecture using a single-visit setting, selecting the most recent visit for each subject. This visit corresponds either to the last confirmed negative exam (validated by a subsequent negative screening) or, for breast cancer cases, to the exam that revealed a suspicious lesion later confirmed by biopsy. The objective of this initial phase was to train a baseline version of the CNN model capable of detecting lesions in a highly imbalanced dataset.

In the second step, we incorporated longitudinal information by including prior exams—-alone or combined with the current visit—-for women with at least five screening examinations. This step aimed to evaluate the model's ability to leverage varying amounts of historical data.

\subsubsection{Step 1: Single-visit training (called \textit{"1Call"})}
\label{sec::met_1Call_training}
In this first step, the model was trained to predict breast cancer using a single screening visit per subject. Specifically, we selected the most recent exam available for each woman, resulting in a cohort of 131,297 women and 525,188 images. A subject-level data splitting strategy was applied, stratified by breast cancer status, to allocate 80\% of the data to training, and 10\% each to validation and test sets \cite{bradshaw_guide_2023}.

To accelerate training and evaluation, we constructed reduced validation and test subsets, each consisting of approximately 4,000 women. All positive cases were retained in these subsets, and a balanced number of negative cases were randomly sampled to maintain a consistent positive-to-negative ratio across both sets. The full training set comprised 105,082 women. To mitigate the pronounced imbalance in outcome distribution, we adopted a sampling strategy that ensured each batch included three negative cases for every positive case. Sampling was conducted at the patient level, and the strategy aimed to enhance the model's exposure to positive cases during training (see Section~\ref{sec::met_model_eval} for additional details).

Model initialization used a pretrained CNN backbone, with gradients propagated through the full architecture, including the GRU and dense layers. We optimized the model using the AdamW algorithm \cite{loshchilov_decoupled_2019} with a weight decay of $10^{-4}$ and a batch size of 8. To identify the optimal training configuration, we compared two fine-tuning strategies: (i) partial fine-tuning, where the CNN backbone was frozen and only the projector, GRU, and dense layers were trained, and (ii) full fine-tuning, where all model components were updated. For each of these, we evaluated two learning rate schemes: a fixed learning rate of $10^{-5}$ and a cosine annealing schedule ranging from an initial value of $10^{-4}$ to a minimum of $10^{-7}$ \cite{loshchilov_decoupled_2019}.

Training progress was monitored using the binary cross-entropy loss \cite{goodfellow_deep_2016} on the validation set. We also tracked the area under the Receiver Operating Characteristic curve (AUC) \cite{metz_roc_1986} to assess model stability and predictive consistency. The model checkpoint yielding the lowest validation loss was selected and evaluated on the test set to confirm generalization performance prior to initiating the second training phase (see Figure~\ref{f:flowchart}).

\subsubsection{Step 2: Incorporating prior examinations (called \textit{"with priors"})}
\label{sec::met_with_prior_training}
Building upon the best-performing model from Step 1, we initialized a second training phase to incorporate longitudinal data. The CNN backbone was frozen to preserve its lesion-detection capability, while the projector, GRU, and dense layers were fine-tuned. We introduced several scenarios based on the number and type of input visits. All included prior exams were confirmed negative. For cancer cases, the current visit (with the visible lesion) was optionally included depending on the scenario.

\begin{figure}[htbp]
 \centerline{\includegraphics[width=6in]{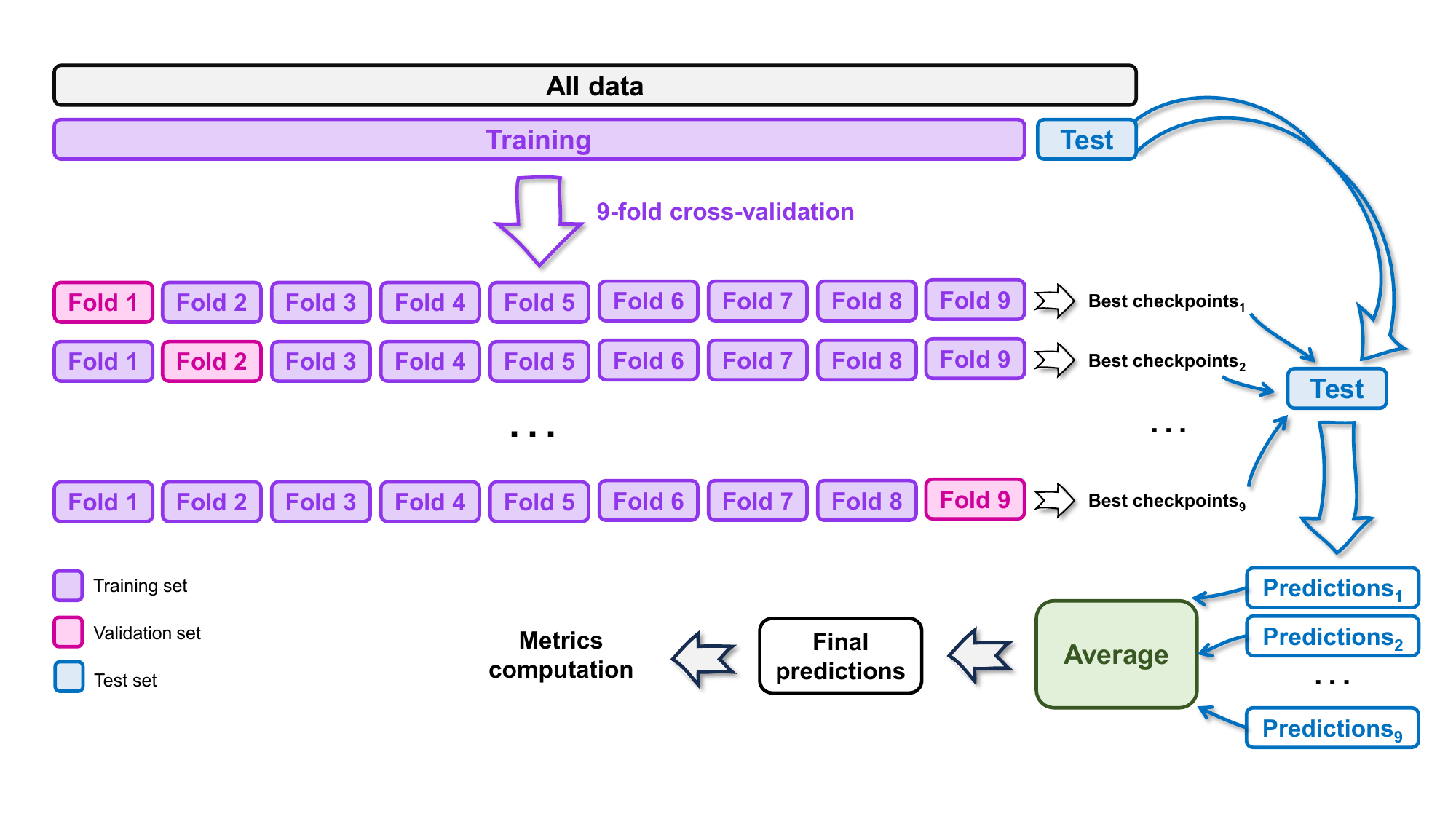}}
\caption{\textbf{Schematic of the 9-fold, holdout test set cross-validation procedure used for model training and evaluation on longitudinal data}. The dataset is divided into a training set and an independent test set. Within the training set, 9-fold cross-validation is performed: in each fold, one partition serves as the validation set while the remaining folds comprise the training data. The best model checkpoints from each fold are used to generate predictions on the test set. Test set predictions are averaged across folds to produce final predictions and compute performance metrics.}
\label{f:crossval}
\end{figure}

After applying the exclusion criteria as illustrated in Figure~\ref{f:flowchart}, the resulting dataset comprised 26,760 women and 533,700 images. We adopted a 9-fold holdout test set cross-validation strategy to robustly assess model generalizability. The full dataset was first split into training and test sets in a 9:1 ratio at the subject level, stratified by breast cancer status. Within the training set, a 9-fold cross-validation procedure was implemented, wherein each fold was sequentially used for validation while the remaining eight folds served for training \cite{bradshaw_guide_2023}. Following each training iteration, the model weights corresponding to the lowest validation loss were preserved. A schematic overview of this cross-validation strategy is provided in Figure~\ref{f:crossval}.

Model optimization was performed using the AdamW optimizer with a fixed learning rate of $10^{-5}$ and a weight decay of $10^{-4}$. We used a batch size of 4, composed of three randomly sampled negative cases and one positive case per batch to address the disproportionate case-to-control ratio.

\subsection{Scenarios}
\label{sec::met_scenarios}
To evaluate the impact of longitudinal information on predictive performance, we defined several scenarios based on the number and type of input visits. These scenarios were grouped into three categories:

\subsubsection{Current visit only}
This group includes a single scenario, denoted as \textbf{1C}, in which only the most recent (current) screening visit is used as input. This configuration serves as a reference, allowing us to evaluate model performance on the same subset of women across all scenarios and to establish a baseline for comparison against settings incorporating prior examinations.

\subsubsection{Priors + current visit}
This group aims to assess the added predictive value of integrating prior exams with the current visit and to explore whether increasing the amount of historical data improves—or potentially impairs—the model’s performance. Four scenarios were defined based on the number of prior visits included alongside the current exam: \textbf{1P1C}: current visit plus the most recent prior, \textbf{2P1C}: current visit plus the two most recent priors, \textbf{3P1C}: current visit plus the three most recent priors, and \textbf{4P1C}: current visit plus four prior visits.

\subsubsection{Priors only}
In this group, the current visit is excluded to isolate the predictive contribution of prior (all negative) exams. This setting allows us to explore whether longitudinal patterns—potentially imperceptible to radiologists—can be leveraged for early risk prediction. We defined four scenarios based on the number of prior exams used as input: \textbf{1P}: most recent prior only, \textbf{2P}: two most recent priors, \textbf{3P}: three most recent priors, and \textbf{4P}: four prior exams (as in Dadsetan et al. (2022) \cite{dadsetan_deep_2022}).

We acknowledge that scenarios with fewer priors (e.g., 1P and 2P) may be suboptimal for sequential modeling due to the limited temporal context available to the GRU. Nonetheless, these configurations were included to assess model robustness under varying levels of longitudinal information.

\subsection{Model evaluation}
\label{sec::met_model_eval}
Model training was performed for a maximum of 40 epochs in both training phases. Early stopping was implemented based on the validation loss, with training interrupted if no improvement greater than $10^{-4}$ was observed over 15 consecutive epochs. Model implementation was carried out using the PyTorch Lightning library (version 2.3.1) \cite{falcon_pytorch_2019}. All experiments were run on a computing node equipped with an Intel CPU (2.40GHz), 128 GB RAM, and multiple NVIDIA Tesla K80 GPUs.

Throughout training, we employed a subject-level batch sampling strategy, where each batch consisted of three randomly selected negative cases and one positive case. This distribution was enforced using a custom sampler that constructed training batches according to a predefined sequence. This sampling strategy was selected to enhance effective learning by exposing the model to a greater diversity of controls while ensuring sufficient representation of positive cases. This sampling scheme was applied exclusively during training. In the validation phase, we corrected for the artificially induced balance in outcome categories by utilizing a weighted loss function, thereby ensuring that evaluation metrics accurately reflect the underlying class distribution.

The binary cross-entropy (BCE) loss function \cite{goodfellow_deep_2016} used is defined as follows:
\begin{equation} \label{eq:binary_cross_entropy}
	\begin{array}{ll}
		l_{c}(x,y) & = \ L_{c} \ = \ \{l_{1,c}, ..., l_{M,c} \}^\top\\
		l_{m,c} & = \ - w_{m,c} \Big{[} \  y_{m,c}\log \sigma(x_{m,c}) + (1 - y_{m,c}) \log (1 - \sigma(x_{m,c}))\Big{]}\\
	\end{array}
\end{equation}
where $M$ represents the batch size, $c$ is the category index ($c=1$ for cancer outcome), and $m$ denotes the $m^{th}$ sample in the batch. The variables $x$ and $y$ represent the model's estimated risk probabilities and ground truth labels, respectively, and $\sigma(x_{m,c})$ corresponds to the sigmoid activation applied to the prediction for sample $m$ and category $c$. During training, no weighting was applied (i.e., $w_{m,c}=1$ for all samples). However, during validation, we applied a correction to account for the sampling bias by setting $w_{m,c}=3$ for negative cases and $w_{m,c}=1$ for positive cases. The final loss was computed by averaging over all samples in the batch.

Model performance was primarily assessed using the AUC on the held-out test set. Ninety-five percent confidence intervals (95\% CIs) were estimated using bootstrap resampling \cite{maier-hein_metrics_2024}. For Step 1 (\textit{"1Call"}), where a simple train/validation/test split was used, AUC was directly computed on the test set predictions. For Step 2, which employed a 9-fold holdout test set cross-validation, we implemented an ensemble prediction strategy: model predictions for the test set were generated using the best checkpoint from each cross-validation fold, and the final test predictions were obtained by averaging across folds. The aggregated predictions were then used to calculate the overall AUC across the nine training folds \cite{bradshaw_guide_2023}. This approach provided a more robust estimate of model performance across the entire training cohort (Figure~\ref{f:crossval}).

\subsection{Subgroup analyses}
\label{sec::met_subgroups}
We evaluated the final model's performance in subgroups defined by two established breast cancer risk factors: mammographic density and age \cite{boyd_mammographic_2007, anderson_comparison_2006}. Three subgroup definitions were applied: First, we stratified women at the current screening visit based on the breast imaging reporting and data system (BI-RADS) mammographic density \cite{spak_bi-rads_2017}, classified as non-dense (BI-RADS categories A and B) or dense (categories C and D), and by age, assigning women to groups younger than 55 years or 55 years and above. The age cutoff of 55 years was selected because it marks the possible transition to biennial screening in the United States and corresponds to an age group with increased breast cancer incidence, when most women have reached menopause \cite{oeffinger_breast_2015, collaborative_group_on_hormonal_factors_in_breast_cancer_menarche_2012, national_cancer_institut___seer_cancer_nodate}. Second, for a longitudinal perspective, we evaluated whether a change in BI-RADS density category occurred within the sequence of input exams for each woman. This density change could occur at any screening visit within the temporal sequence, not necessarily at the current visit (For instance, between prior 3 and 2). As a result, the estimated density change depended on the considered scenario.
\section{Results}
\label{sec::results}

\subsection{Patient characteristics}
\label{sec::patient_char}
Table~\ref{t:Women_chars} presents the demographic and clinical characteristics of the study population at the most recent (current) screening visit according to study phase (\textit{"1Call"} and \textit{"with priors"}) and breast cancer status. Screening period extended from 2006 to 2019. In the \textit{"1Call"} step, the dataset included $131{,}297$ women, of whom $3{,}129 \ (2.38\%)$ were diagnosed with breast cancer. In the \textit{"with priors"} step, $26{,}760$ women met the longitudinal inclusion criteria, including $678 \ (2.53\%)$ cancer cases. Women with cancer were generally older than cancer-free participants across both study phases. The distribution of participants across centers and imaging manufacturers was slightly imbalanced, particularly in Step 2, where the majority of women were screened at Center 1 and imaged with Hologic equipment. Regarding mammographic density, cancer cases tended to present with higher mammographic density (C or D) compared to cancer-free women. For the Step 2 cohort, median intervals between successive prior visits were consistent, typically around one year, reflecting adherence to routine American screening schedules.

\begin{table}[htbp]
\caption{\textbf{Key patient characteristics.} Summary of patient characteristics assessed at the most recent (current) screening visit for the cohorts included in Step 1 (\textit{"1Call"}) and Step 2 (\textit{"with priors"}) over the screening period from 2006 to 2019.}
\label{t:Women_chars}
\begin{center}
\resizebox{\textwidth}{!}{%
\renewcommand{\arraystretch}{1.1}
\begin{tabular}{lccc|ccc}
\hline
\multirow{4}{*}{\textbf{Variable}} & \multicolumn{3}{c}{\textbf{Step 1: Single visit training (\textit{"1Call"})}} & \multicolumn{3}{c}{\textbf{Step 2: Incorporating prior examinations (\textit{"with priors"})}} \\
\cline{2-7}
 & \multicolumn{2}{c}{\textbf{Breast cancer status}} & \textbf{Total} & \multicolumn{2}{c}{\textbf{Breast cancer status}} & \textbf{Total} \\
 & Cancer-free & Cancer &  & Cancer-free & Cancer & \\
  & N = 128,168 & N = 3,129 & N = 131,297 & N = 26,082 & N = 678 & N = 26,760 \\
\hline
\textbf{Age at the current visit, y} & & & & & & \\
& 55 [47--65]  & 61 [51--69] & 55 [47--65] & 60 [52--67]  & 63 [54--69] & 60 [52--68] \\
\textbf{Center} & & & & & & \\
 \quad 1 & 78,748 (61.4) & 1,894 (60.5) & 80,642 (61.4)  & 22,472 (86.2) & 594 (87.6) & 23,066 (86.2) \\
 \quad 2 & 49,420 (38.6) & 1,235 (39.5) & 50,655 (38.6)  & 3,610 (13.8) & 84 (12.4) & 3,694 (13.8) \\
\textbf{Manufacturer} & & & & & & \\
 \quad Hologic & 76,703 (59.8) & 1,858 (59.4) & 78,561 (59.8) & 16,674 (63.9) & 393 (58.0) & 17,067 (63.8) \\
 \quad GEHC & 51,465 (40.2) & 1,271 (40.6) & 52,736 (40.2)  & 9,408 (36.1) & 285 (42.0) & 9,693 (36.2) \\
\textbf{BI-RADS score} & & & & & & \\
 \quad A & 18,679 (14.6) & 296 (9.5) & 18,975 (14.5)  & 4,256 (16.3) & 44 (6.5) & 4,300 (16.1) \\
 \quad B & 47,089 (36.7) & 1,169 (37.4) & 48,258 (36.7) & 10,480 (40.2) & 260 (38.3) & 10,740 (40.1) \\
 \quad C & 51,442 (40.1) & 1,431 (45.7) & 52,873 (40.3) & 9,767 (37.4) & 330 (48.7) & 10,097 (37.7) \\
 \quad D & 10,958 (8.6) & 233 (7.4) & 11,191 (8.5) & 1,579 (6.1) & 44 (6.5) & 1,623 (6.1) \\
\textbf{Duration in years between} & & & & & & \\
 \quad 1st prior and current visit & \multicolumn{3}{c|}{\cellcolor{gray!10!white}} & 1.06 [1.02--1.86]  & 1.05 [1.01--1.38] & 1.06 [1.02--1.84] \\
 \quad 2nd prior and 1st prior & \multicolumn{3}{c|}{\cellcolor{gray!10!white}} & 1.04 [1.01--1.25]  & 1.04 [1.01--1.23] & 1.04 [1.01--1.25] \\
 \quad 3rd prior and 2nd prior & \multicolumn{3}{c|}{\cellcolor{gray!10!white}} & 1.04 [1.01--1.21]  & 1.04 [1.01--1.21] & 1.04 [1.01--1.21] \\
 \quad 4th prior and 3rd prior & \multicolumn{3}{c|}{\cellcolor{gray!10!white}} & 1.05 [1.01--1.22]   & 1.05 [1.01--1.23]  & 1.05 [1.01--1.22] \\
\hline
\multicolumn{7}{l}{Data are median [IQR] and n (\%)}\\
\multicolumn{7}{l}{GEHC: General Electric Healthcare; BI-RADS: breast imaging reporting and data system}\\
\end{tabular}%
}
\end{center}
\end{table}

\begin{figure}[htbp]
 \centerline{\includegraphics[width=3.8in]{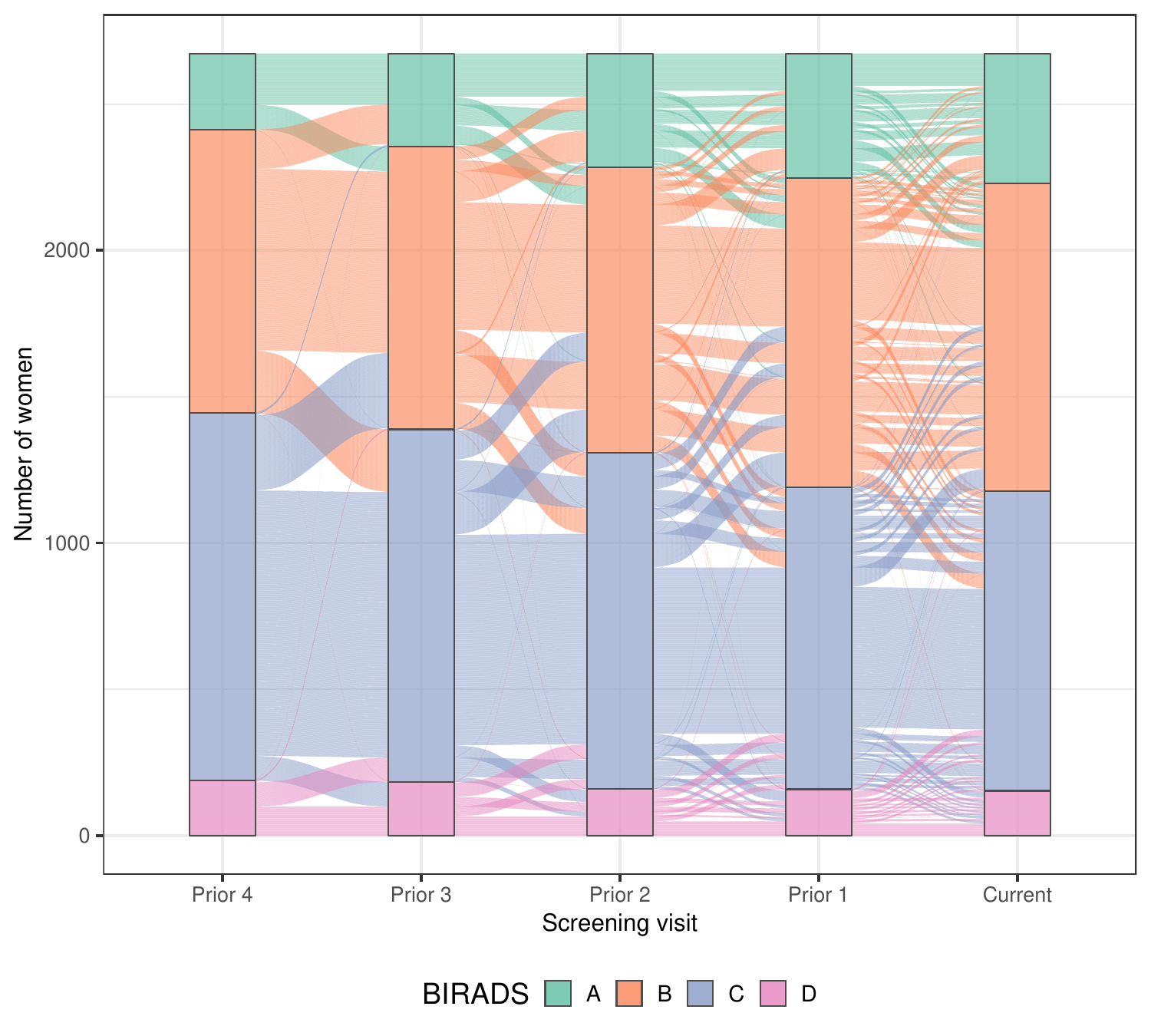}}
\caption{\textbf{Longitudinal evolution of BI-RADS breast density categories over five screening visits.} The chart presents the distribution of women across BI-RADS density categories (A to D) from the most distant prior visit (Prior 4) to the current visit. Colored segments indicate the proportion of women within each density category at each visit. Connecting lines represent individual transitions between categories across visits, illustrating changes in breast density over time.}
\label{f:birads_evolution}
\end{figure}

Figure~\ref{f:birads_evolution} shows the evolution of BI-RADS categories across five sequential screening visits, including the current and up to four prior exams. Most women remained in the same BI-RADS category over time, particularly in categories B and C. The figure also highlights a subgroup of women who experienced shifts in mammographic density, with an overall increase in the proportion of non-dense classes (A and B) and a decrease in dense classes (C and D). This pattern reflects the dynamic nature of breast tissue and the expected decline in mammographic density over time, consistent with previous research \cite{burton_mammographic_2017}. In addition, some women showed changes toward higher density between two adjacent BI-RADS categories (e.g., from B to C). Such increases can arise from two main causes: either a true but uncommon rise in density, or variability in the subjective classification of BI-RADS, especially in women whose mammograms are difficult to categorize clearly. This phenomenon has been previously reported and reflects a known limitation of BI-RADS density assessment \cite{portnow_persistent_2022}.

\subsection{\textit{"1Call"} training}
\label{sec::res_1call}
Table~\ref{t:res_1call} presents the performance of the four candidate training strategies explored during the single-visit training phase (\textit{"1Call"}), where models were trained using only the most recent (current) screening examination. Among the four configurations, the best performance was achieved by the model trained with full fine-tuning and a fixed learning rate, reaching an AUC of 0.742 (95\% CI: 0.711--0.773). This result outperformed the same fine-tuning strategy with an adaptive learning rate, which yielded an AUC of 0.710 (0.676--0.740), suggesting that a conservative fixed learning rate was more effective in this context. The two strategies based on partial fine-tuning, whether with a fixed (0.70~(0.672--0.732)) or adaptive learning rate (0.698 (0.665--0.729)), showed consistently lower performance, indicating limited benefit when the CNN backbone remained frozen. Based on these findings, the model trained with full fine-tuning and fixed learning rate was selected for the subsequent stage of the study involving longitudinal inputs.

\begin{table}[htbp]
\caption{\textbf{Test set model performance during the \textit{"1Call"} step ($N = 4{,}216$ women).} Area under the ROC curve (AUC) values with corresponding $95\%$ confidence intervals ($95\%$ CI), comparing four training configurations based on the extent of fine-tuning (partial vs. full) and the learning rate strategy (fixed vs. adaptive).}
\label{t:res_1call}
\begin{center}
\begin{small}
\renewcommand{\arraystretch}{1.1}
\begin{tabular}{llc}
\hline
\multicolumn{3}{c}{\textbf{$N = 4{,}216$ women}}\\
\hline
\textbf{Fine-tuning} & \textbf{Learning rate} & \textbf{AUC ($95\%CI$)}\\
\hline
\multirow{2}{*}{Partial} & Fixed ($10^{-5}$) & 0.701 (0.672--0.732) \\
 & Adaptive* & 0.698 (0.665--0.728) \\
\hline
\multirow{2}{*}{Full} & Fixed ($10^{-5}$) & \textbf{0.742 (0.711--0.773)} \\
 & Adaptive* & 0.710 (0.676--0.740) \\
\hline
\multicolumn{3}{l}{*: Cosine annealing schedule (from $10^{-4}$ to $10^{-7}$) }\\
\multicolumn{3}{l}{Bold AUC ($95\%CI$) represents the best performing model} \\
\multicolumn{3}{l}{$95\%$ CI were computed via bootstrapping}
\end{tabular}
\end{small}
\end{center}
\end{table}

\subsection{Training \textit{"with priors"}}
\label{sec::res_priors}
Table~\ref{t:res_priors} presents the model’s predictive performance across the three input scenario groups. In the \textit{Current visit only} group, which used only the most recent screening exam, the model achieved an AUC of 0.767 (0.702–0.829), indicating good discriminative ability from a single time point.

In the \textit{Priors + current visit} group, combining the current exam with up to four prior negative exams, performance remained stable, with AUCs ranging from 0.764 to 0.770. The highest value was observed with three prior exams (3P1C: AUC = 0.770 (0.709–0.834)). These results suggest that including prior images alongside the current exam offers a slight performance gain, with a possible optimal balance at three priors.

In the \textit{Priors only} group, where the current exam was excluded, performance decreased across all configurations. The best result was obtained with two prior exams (2P: AUC = 0.674 (0.606–0.740)). Our dataset, characterized by heterogeneous time intervals between visits and a pronounced imbalance between cancer and cancer-free outcomes, posed greater challenges but also better reflected real-world screening practice. Despite these difficulties, performance in this scenario group remained comparable to that reported by Dadsetan et al. (2022) \cite{dadsetan_deep_2022}, the only configurations also evaluated in their study. The lower values compared to \textit{Current visit only} were expected, as the current exam likely contains the most relevant lesion-specific information.

Overall, the results indicate that the current image carries the most important information for accurate breast cancer prediction, while prior exams can provide incremental benefit when combined with it. In contrast, relying solely on prior negative exams limits predictive ability. In the absence of the current exam, the model likely relies on non-specific lesion texture patterns that may be present before diagnosis but remain undetectable during routine visual assessment by radiologists.

\begin{table}[htbp]
\caption{\textbf{Test set model performance during the \textit{\textit{"with priors"}} step ($N = 2{,}673$ women).} Area under the ROC curve (AUC) values with corresponding $95\%$ confidence intervals ($95\% CI$), comparing the model’s breast cancer predictive performance across three scenario groups: \textit{Current visit only} (most recent exam only), \textit{Priors + current visit} (current exam combined with up to four prior negative exams), and \textit{Priors only} (up to four prior negative exams, excluding the current visit).}
\label{t:res_priors}
\begin{center}
\begin{small}
\renewcommand{\arraystretch}{1.1}
\begin{tabular}{lc}
\hline
\multicolumn{2}{c}{\textbf{$N = 2{,}673$ women}}\\
\hline
\textbf{Scenarios} & \textbf{AUC ($95\%CI$)} \\
\hline
\textbf{Current visit only} & \\
 \quad 1C & 0.767 (0.702--0.829) \\
\hline
\textbf{Priors + current visit} & \\
 \quad 1P1C & 0.768 (0.704--0.831) \\
 \quad 2P1C & 0.768 (0.702--0.830) \\
 \quad 3P1C & \textbf{0.770 (0.709--0.834)} \\
 \quad 4P1C & 0.764 (0.701--0.820) \\
\hline
\textbf{Priors only} & \\
 \quad 1P & 0.644 (0.572--0.714) \\
 \quad 2P & \textbf{0.674 (0.606--0.740)} \\
 \quad 3P & 0.664 (0.594--0.732) \\
 \quad 4P & 0.659 (0.591--0.726) \\
\hline
\multicolumn{2}{l}{Bold values represent the best performing model in the corresponding group} \\
\multicolumn{2}{l}{$95\%$ CI were computed via bootstrapping}
\end{tabular}
\end{small}
\end{center}
\end{table}

Regarding subgroup analyses, results are presented in Table~\ref{t:res_priors_2} and~\ref{t:res_priors_3}. When stratified by mammographic density, the model performed consistently across scenarios, with higher AUC values in non-dense than in dense breasts. This difference likely reflects that lower-density images provide clearer visual cues for malignancy. In the \textit{Priors + current visit} scenarios, performance improved compared to the \textit{Current visit only} scenario, indicating that prior examinations add predictive information even in dense tissue. In contrast, the \textit{Priors only} scenarios showed reduced performance relative to 1C, suggesting that the absence of the current exam limits the model’s ability to capture relevant features, particularly in dense breasts.

When stratified by age, AUC values were consistently higher in women younger than 55 years than in those aged 55 years or older. In both age groups, combining prior and current examinations yielded higher AUC values than using the current visit alone. As with density, the \textit{Priors only} scenarios showed lower performance than scenarios that included the current exam, which again highlights the importance of mammograms from the most recent visit for prediction. Although performance in the older group was reduced across all scenarios, AUC values remained relatively high, indicating that the model retains discriminative ability in the age group where breast cancer occurs most frequently.

\begin{table}[htbp]
\caption{\textbf{Test set model performance by mammographic density and age across input scenarios during the \textit{"with priors"} step ($N = 2{,}673$ women).} Area under the ROC curve (AUC) values with corresponding $95\%$ confidence intervals ($95\%$ CI), comparing model performance according to mammographic density category (non-dense vs. dense) and age ($<$ 55 years vs. 55 years \& above) across three scenario groups: \textit{Current visit only} (most recent examination only), \textit{Priors + current visit} (current examination combined with up to four prior negative examinations), and \textit{Priors only} (up to four prior negative examinations, excluding the current visit).}
\label{t:res_priors_2}
\begin{center}
\resizebox{\textwidth}{!}{%
\renewcommand{\arraystretch}{1.1}
\begin{tabular}{l|cc|cc}
\hline
\multicolumn{5}{c}{\textbf{$N = 2{,}673$ women}}\\
\hline
\multirow{3}{*}{\textbf{Scenarios}} & \multicolumn{4}{c}{\textbf{Subgroups}}\\
\cline{2-5}
 & \multicolumn{2}{c|}{\textbf{Mammographic density}} &
 \multicolumn{2}{c}{\textbf{Age}} \\
\cline{2-5}
 & Non-dense & Dense &  $<$ 55 years & 55 years \& above \\
\hline
\textbf{Current visit only} & & & & \\
\quad 1C & 0.822 (0.733--0.905) & 0.706 (0.606--0.797) & 0.828 (0.716--0.926) & 0.748 (0.666--0.822) \\
\hline
\textbf{Priors + current visit} & & & & \\
\quad 1P1C & 0.803 (0.710--0.891) & 0.717 (0.626--0.804) & 0.831 (0.715, 0.928) & 0.748 (0.667, 0.818) \\
\quad 2P1C & 0.811 (0.717--0.896) & 0.714 (0.612--0.808) & 0.831 (0.727, 0.920) & 0.747 (0.662, 0.822) \\
\quad 3P1C & 0.796 (0.697--0.887) & \textbf{0.723 (0.627--0.814)} & \textbf{0.836 (0.733, 0.925)} & \textbf{0.749 (0.669, 0.822)} \\
\quad 4P1C & \textbf{0.820 (0.738--0.899)} & 0.707 (0.604--0.802) & 0.813 (0.699, 0.908) & 0.747 (0.667, 0.819) \\
\hline
\textbf{Priors only} & & & & \\
\quad 1P & 0.624 (0.509--0.746) & 0.615 (0.525--0.705) & 0.691 (0.570, 0.800) & 0.630 (0.549, 0.715) \\
\quad 2P & 0.659 (0.545--0.770) & 0.641 (0.553--0.730) & 0.703 (0.578, 0.815) & \textbf{0.664 (0.589, 0.748)} \\
\quad 3P & \textbf{0.661 (0.548--0.778)} & \textbf{0.626 (0.532--0.722)} & \textbf{0.707 (0.589, 0.824)} & 0.650 (0.570, 0.736) \\
\quad 4P & 0.645 (0.530--0.758) & \textbf{0.626 (0.535--0.716)} & 0.691 (0.572, 0.809) & 0.648 (0.570, 0.731) \\
\hline
\multicolumn{5}{l}{Mammographic density and age groups are evaluated at the current visit.} \\
\multicolumn{5}{l}{Non-dense ($N = 1{,}496$ women) vs. dense ($N = 1{,}177$ women)} \\
\multicolumn{5}{l}{$<$ 55 years ($N = 689$ women) vs. 55 years \& above ($N = 1{,}984$ women)} \\
\multicolumn{5}{l}{Bold values represent the best performing model in the corresponding group}\\
\multicolumn{5}{l}{$95\%$ CI were computed via bootstrapping}
\end{tabular}%
}
\end{center}
\end{table}

For the analysis by change in mammographic density (Table~\ref{t:res_priors_3}), the model performed better in the change group than in the no change group in all scenarios. The gain in AUC ranged from +0.062 to +0.138 in the \textit{Priors only} group and from +0.067 to +0.112 in the \textit{Priors + current visit} group. This improvement was consistent across both prior-limited scenarios (1P, 1P1C) and prior-enriched scenarios (4P, 4P1C). These results indicate that texture changes in mammographic density provide useful predictive information for the model and highlight the benefit of including prior images to refine breast cancer risk prediction in specific situations.

\begin{table}[htbp]
\caption{\textbf{Test set model performance by change in mammographic density across input scenarios during the \textit{"with priors"} step ($N = 2{,}673$ women).} Area under the ROC curve (AUC) values with corresponding $95\%$ confidence intervals ($95\%$ CI), comparing model performance across three scenario groups: \textit{Current visit only} (most recent examination only), \textit{Priors + current visit} (current examination combined with up to four prior negative examinations), and \textit{Priors only} (up to four prior negative examinations, excluding the current visit).}
\label{t:res_priors_3}
\begin{center}
\resizebox{\textwidth}{!}{%
\renewcommand{\arraystretch}{1.1}
\begin{tabular}{clc|clc}
\hline
\multicolumn{6}{c}{\textbf{$N = 2{,}673$ women}}\\
\hline
\multicolumn{3}{c|}{\textbf{Priors + current visit}} & \multicolumn{3}{c}{\textbf{Priors only}} \\
\hline
\textbf{Scenario} & \textbf{Change in density} & \textbf{AUC ($95\%$ CI)} & \textbf{Scenario} & \textbf{Change in density} & \textbf{AUC ($95\%$ CI)} \\
\hline
\multirow{2}{*}{1P1C} & No change ($N = 1{,}927$) & 0.727 (0.632--0.812) &\multirow{2}{*}{1P} & No change ($N = 1{,}927$) & 0.603 (0.512--0.699)\\
& Change ($N = 746$) & \textbf{0.822 (0.746--0.893)} & & Change ($N = 746$) & \textbf{0.692 (0.595--0.782)} \\
\hline
\multirow{2}{*}{2P1C} & No change ($N = 1{,}455$) & 0.725 (0.613--0.819) &\multirow{2}{*}{2P} & No change ($N = 1{,}455$) & 0.600 (0.505--0.696)\\
& Change ($N = 1{,}218$) & \textbf{0.805 (0.715--0.877)} & & Change ($N = 1{,}218$) & \textbf{0.738 (0.643--0.813)} \\
\hline
\multirow{2}{*}{3P1C} & No change ($N = 1{,}146$) & 0.701 (0.582--0.822) &\multirow{2}{*}{3P} & No change ($N = 1{,}146$) & 0.603 (0.502--0.709)\\
& Change ($N = 1{,}527$) & \textbf{0.813 (0.743--0.877)} & & Change ($N = 1{,}527$) & \textbf{0.701 (0.609--0.784)} \\
\hline
\multirow{2}{*}{4P1C} & No change ($N = 1{,}927$) & 0.717 (0.579--0.840) &\multirow{2}{*}{4P} & No change ($N = 1{,}927$) & 0.617 (0.517--0.713) \\
& Change ($N = 746$) & \textbf{0.784 (0.713--0.855)} & & Change ($N = 746$) & \textbf{0.679 (0.590--0.763)} \\
\hline
\multicolumn{6}{l}{$N$ represents number of women}\\
\multicolumn{6}{l}{Bold values represent the best performing model in the corresponding group}\\
\multicolumn{6}{l}{$95\%$ CI were computed via bootstrapping}
\end{tabular}%
}
\end{center}
\end{table}

\section{Discussion}
\label{sec::discussion}
In this study, we introduced the LongiMam model, a new end-to-end framework for breast cancer risk prediction from single and longitudinal mammograms, built on a previously proposed approach \cite{dadsetan_deep_2022}. The model was developed with large, real-world data from two screening centers, reflecting the imbalance between cancer and cancer-free women that characterizes population-based screening. To address this challenge, we applied tailored training strategies and released the approach as an open-source solution on \href{https://github.com/manelrakez/LongiMam.git}{GitHub}. LongiMam integrates the most recent exam with up to four prior mammograms to estimate individual risk at the current visit.

Model training followed a two-step process. First, we trained the CNN backbone for breast cancer risk prediction using the most recent exam only. Next, we extended the model to longitudinal data, testing different input sequences with varying numbers and types of exams. In addition to a \textit{priors only} configuration, previously studied by Dadsetan et al. (2022) \cite{dadsetan_deep_2022}, we evaluated scenarios that included both current and prior exams (\textit{priors + current visit}) to understand each data type’s contribution.

LongiMam achieved robust performance across all examined scenarios. In the \textit{current visit only} setting, the model reached an AUC of 0.767 (0.702–0.829), discriminating well between cases and controls despite the low prevalence of cancer ($2.53\%$). When prior mammograms were added, performance improved in most settings, indicating that prior exams contribute additional useful information without impairing accuracy. However, a plateau effect was observed, as performance did not improve further when more than three priors were included, which may reflect the limited contribution of distant priors. This observation aligns with the concept of sojourn time, the preclinical period during which cancer can be detected by screening. Previous research estimates this period at about 2.8 years for women aged 50–69 \cite{weedon-fekjaer_breast_2008}, which aligns with the median age of 60 in our cohort. Within this timeframe, recent mammograms may capture most of the predictive information, while older exams add little. Screening sensitivity also increases with age \cite{cong_estimation_2005}, which may further support detectability in an older population. Together, these findings can provide a biological and clinical rationale for prioritizing recent and temporally relevant prior imaging when developing predictive models.

In the \textit{priors only} scenarios, the model achieved AUC values between 0.644 and 0.674, close to those reported in the original study based on a much smaller dataset of 200 women with matched case-control sampling \cite{dadsetan_deep_2022}. These findings suggest that our modifications and training strategy allowed the model to adapt to a more complex and heterogeneous setting. A direct comparison with LRP-NET was not feasible, as model weights were unavailable and the datasets differed substantially. In particular, Dadsetan et al. (2022) \cite{dadsetan_deep_2022} used a small, artificially balanced dataset with a density distribution that does not represent screening populations, including an under-representation of extremely dense (class D) breasts.

The addition of prior exams consistently improved performance when combined with the current visit across analysed subgroups. By contrast, when prediction relied only on prior negative exams, performance decreased relative to the \textit{current visit only} setting but remained acceptable. Importantly, the model retained good discriminative ability in two challenging groups: women with dense breasts and women aged 55 years or older. Although a direct comparison was not possible, these findings are consistent with previous studies that evaluated similar subgroups with other AI models on diverse datasets \cite{lee_enhancing_2023, damiani_evaluation_2023}.

We further assessed changes in mammographic density over time. The model performed better in women who experienced a shift in BIRADS category than in those with stable density, across all scenarios. This supports the use of prior exams to capture temporal changes that may better inform on risk evolution. Our findings align with previous reports \cite{lee_enhancing_2023} and can suggest that longitudinal changes in density are predictive of breast cancer \cite{jiang_longitudinal_2023}.

A major strength of this study is the use of a large, real-world dataset. We did not restrict the time interval between visits, exposing the model to a diverse spectrum of real-world screening practices. To ensure clear evaluation, we applied scenarios where all women had the same number of visits, but this approach does not fully capture the variability found in routine care. Adapting future model architectures to handle sequences of varying length and irregular intervals should improve clinical applicability \cite{rubanova_latent_2019}.

Our dataset, though substantial, was collected from only two centers, which may limit the model’s generalizability. External validation on independent cohorts is therefore an essential next step to test robustness and reliability across diverse populations and clinical environments.

Few earlier studies have assessed the impact of integrating prior mammograms in risk modeling \cite{dadsetan_deep_2022}. While the added benefit of priors was modest when the current exam was available, the LongiMam model shows promise as a decision support tool for radiologists \cite{roelofs_importance_2007}. Most risk prediction methods focus on short- or long-term estimation without full use of longitudinal information \cite{lee_enhancing_2023, yala_toward_2021, donnelly_asymmirai_2024, yeoh_radifusion_2023, yeoh_new_2024, karaman_longitudinal_2024}. A relevant extension would be to replace the GRU and dense layers with survival-oriented modules to enable dynamic risk prediction aligned with longitudinal screening data \cite{moon_survlatent_2022}.

Another perspective is to examine the interpretability and explainability of the model. Gradient-based attribution methods, such as saliency maps \cite{simonyan_deep_2014} and Grad-CAM \cite{selvaraju_grad-cam_2020}, can generate visual explanations that indicate which regions of the image contribute most to the risk prediction. Such outputs can improve trust, support clinical decision-making, and reveal imaging patterns that might otherwise remain unnoticed. Moreover, interpretability is essential for identifying potential biases and ensuring consistent performance across subgroups of women \cite{klanecek_longitudinal_2024}.

In summary, the LongiMam model demonstrates solid performance and flexibility for breast cancer risk prediction in real-world screening, particularly when both current and prior mammograms are available. The model performed well in subgroups of women aged 55 years and older, those with dense breasts, and those who experienced changes in breast density over time. With further external validation and advancements toward dynamic risk modeling and explainability, LongiMam has the potential to enhance personalized screening and support clinical decision-making.

\section*{Funding}
\label{sec::funding}
This work was supported by the "Institut National du Cancer" [grant number: INCa\textunderscore16049]; "La Ligue de la Gironde et des Landes"; and the PIA3- Investment for the Future [grant number: 17-EURE-0019].

\bibliographystyle{elsarticle-num}
\bibliography{LongiMam.bib}

\end{document}